# Bipedal Model Based on Human Gait Pattern Parameters for Sagittal Plane Movement

V.B.Semwal, S.A.Katiyar, P.Chakraborty and G.C.Nandi
*Department of Robotics and AI Engineering*
*IIIT Allahabad*
*Allahabad, U.P.-211012, India*
{vsemwal, shiv.ashu.lko, pavan.chakraborty & gcnandi}@gmail.com

*Abstract* - The present research as described in this paper tries to impart how imitation based learning for behavior-based programming can be used to teach the robot. The simulated model tries to imitate human GAIT pattern and negotiate push with efficient recovery [1]. This paper also proposes the HOAP2 [2] based biped model to achieve gait cycle imitation and push recovery on humanoid. The proposed model follows the Gait cycle [1] and can be further used for developing a model capable to recover from push similar to human biology. This development is a big step in way to prove that push recovery is a software engineering problem and not hardware engineering problem. The walking algorithm used here aims to select a subset of push recovery problem i.e. disturbance from environment. We applied the physics at each joint of Halo with some degree of freedom. The proposed model, Halo is different from other models as previously developed model were inconsistent with data for different persons. This would lead to development of the generalized biped model in future and will bridge the gap between performance and inconsistency. In this paper the proposed model is applied to data of different persons. Accuracy of model, performance and result is measured using the behavior negotiation capability of model developed. In order to improve the performance, proposed model gives the freedom to handle each joint independently based on the belongingness value for each joint. The development can be considered as important development for future world of robotics. The accuracy of model is 70% in one go. In this paper, we achieve to imitate the human gait cycle for HOAP-2 [2] robots model Halo. We validate our model by giving different input configuration parameter i.e. CoM, CoP and joint angle of different samples to HOAP-2[2] model designed in Webots, which can demonstrate the behavior as per new configuration provided for different person.

*Index Terms - Push Recovery, HOAP2 robot, Imitate, Artificial Intelligence, GAIT cycle, Hierarchical fuzzy system, Degree of freedom, Human motion capturing device (HMCD [3]).*

## I. INTRODUCTION

Push recovery is the behaviour shown by any subject towards recovery from unexpected external push. Human Push recovery capability is a learning process which human being learns through a continuous evolution process. Our two most sophisticated parts of central nervous system, right lobe and left lobe are used to control and recover from particular push. So it is one of the most frequent and quick decision taken by humans during normal walking. The problem occurs when robot is made to work in similar environments like humans. Using LIPM [4] based biped model which is inherently unstable and has serious stability problem but due to better adaptability, obstacle negotiation capability and stairs climbing capability we are using biped model. We as human prefer to learn through continuous learning by negotiating with environment and by effectively using different strategy to impart human learning robot so we are using fuzzy logic based controller which counters impreciseness of environment.

HOAP2 [2] model is ubiquitous to impart human learning to a biped model [3]. The reason for being so commonly used is because of properties such as mass, density, inertia, learning from examples, physics we can defined and exhibit some capability for generalization beyond the sample data [6]. Also it has universal approximation property [7]. Fuzzy set and logic theory [5] is one of the most prominent tools to handle uncertainty in decision-making. The major benefit of using fuzzy model is to achieve low cost and more intuitive model. Fuzzy model as a tool is easily understandable by human i.e. it has sort of knowledge representation for better human understandability. Obviously, the fuzzy logic is a tool which deals linguistic information [9]. Pattern Classification problem can we solved using two most widely techniques learning and generalization [10].

A biped robot maintains the parameter analogous to human body. To impart learning for push recovery it follows the process like human body follow. Our model Halo, whose architecture is based on HOAP2[1] has twenty-five joints of body in all including head with two degree of freedom, left and right arm with five DoF, two legs with six DoF[11]. This imitates the human gait pattern for its implementation as biped [1]. We distribute the mass to all parts, apply the torque and inertia on each joint to balance model. The paper uniquely identifies the parameters for balancing. To imitate the human like behavior on humanoid is very complex thus to exhibit the human like behavior sensors on 6 joints are used to capture the joint angle and corresponding force sensors, which are used to measure the magnitude of applied force [18].

### A. Paper Organization:

The first section covers the requirement and important with respect to other available model. Then second section contains a brief discussion of the Fujitsu's humanoid open architecture platform-2 (HOAP-2) robot [1]. The section three cover the step of implementation and depth understanding of our model presented, generalization parameter of some properties such as mass, density, inertial etc. physics exhibiting some capability toward development of generalized architecture of robot for push recovery and revealed approach of the walking humanoid robot in their stance and swing phase of human gait cycle [1][17]. Subsequently, we have presented the model in Webots and defined parameter and physics applied and apply smart hierarchical fuzzy logic based controller which covering the degree of membership or non membership using intuitive fuzzy set in section three and four.

In the result and discussion section we have shown the simulate model based on collected experimental data for various physical change corresponding to push. Finally we concluded the push recovery not only depend push recovery not only depends on the person's height, weight or sex but also depends on whether a person is a left or right handed [18].These simulations are a big leap in the further investigation and development step of real biped humanoid robot which will be synergistically similar with the human[19]. Some other relevant and recent works in the area of humanoid push recovery can be obtained in [22].

II. HUMANOID OPEN ARCHITECTURE PLATFORM ROBOT

The HOAP-2[1] is one of the biped robots of HOAP series, developed by Fujitsu for commercial use. It is very much popular between research community due to resemblance with the human with more DoF and mobility. The architectural details are given in manual [1]. The DoF which is actually defining the number of parameter to defined the all joint configuration which help us to analyze the problem of push recovery.

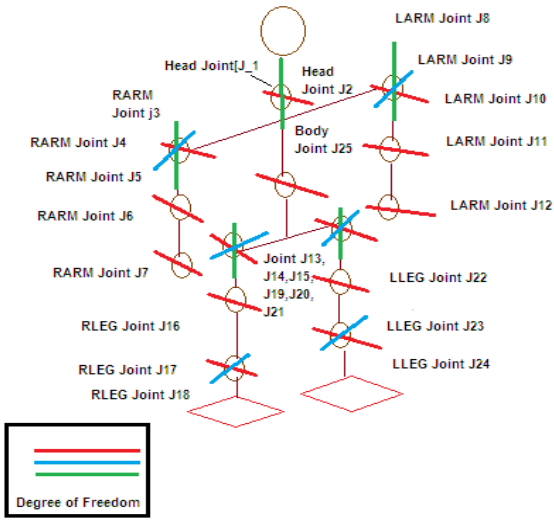

Fig 1: Degree of freedom of Halo

II. OUR APPROACH

A. The Step of methodology

The objective of this paper is to describe the model which can perform the human like Gait cycle and analyze the behavior using hierarchical fuzzy algorithm, control the balance of lateral plane movements [14] of humanoid robots. We taken consideration of the HOAP2 [2] is model because it is resemble to human. We organized our methodology in the following steps:

Step1:- To simulate the liner inverted pendulum equivalent biped model i.e. HOAP2 [1] similar model in Webots. And calibrate the all the part and joint with their respective physical parameter like degree of freedom, mass, centre of mass, height, length of body part etc.

Step2:- To compute the control torques for the joints in a bipedal humanoid, change in angle of different joint, gait pattern taken from HMCD [3] is of human and finally calculate Inverse kinematic and D-H parameter of model.

Step3:-Apply all the above calculated parameter to our model i.e. Halo and make it to follow the human gait cycle based on real time data captured through HMCD [3].

Step4:-To gather the sufficient amount of data by our model to impart the learning technique.

Step5:- To develop a hierarchical fuzzy module [5] based on the data gather through indigenous developed suit HMCD [3] and associate same computing on Webots controller to guide the our model similar to HOAP2 [1] to follow the gait cycle and predict the output angle and velocity on the basis of input force and direction around our rule defined in paper.

Biped which is human resembles inverted pendulum structure is inherently unstable and while interacting with human will ultimately collide with humans, things and other humanoids, So humanoid push and fall recovery is very important to make them absorb force and stand up after falling. In this paper we tried to address the issue of push recovery with referenced to the humanoid robot HOAP2 [1]. The experimental data collected through in house developed HMCD device used in development of software based model. Then to make it follow the human gait cycle, Inverse kinematic have been solved for its legs [19], the model also has been programmed to be visible dynamically. To make advance we involved the two infrared sensors on the eyes of biped to identify obstacle in the way. To make it free from intensive computation mechanism and impart learning we proposed the hierarchical fuzzy system [5].gait cycle style in which a human being walks. Gait cycle starts where one foot leaves the ground and end it again touches the ground [12].

The control torques can be computed for the joints in a bipedal humanoid using the following equations:

$$\tau = M(\theta)\ddot{\theta} + C(\theta,\dot{\theta}) + G(\theta) \quad (1)$$

For normal walking pattern, torques can be generated as in eqn.(1) which is sum of inertial torque (M), Centrifugal and

Coriolis torque(C) and gravitational torque (G) together with some frictional torques which has been neglected. When external force is applied on human, these ideal walking patterns of different joint torques get disturbed. To regain original torques, control torques required to be applied [8].

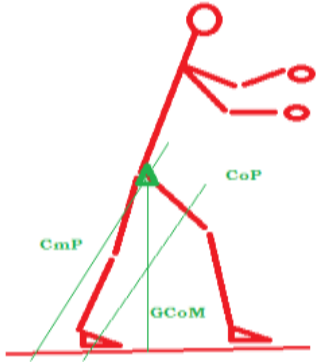

Fig2: Displaying the balance of CmP

Method resolving the present problem

Model Architecture

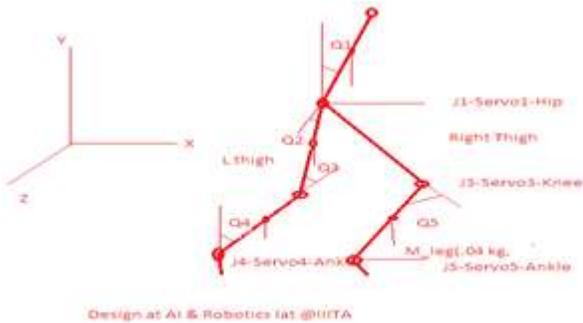

Fig3: Details Design of Model

Our model consists of a sphere surmounted on a cylinder attached with two limbs. Each Limb is equipped with total 4 servos, 2 servos at keen and 2 servo at ankle. The servo which is used at foot does not play any vital role to move the leg at any position but do keep orientation of the feel almost parallel to the ground in swing phase and stance phase of gait cycle.

Inducing anthropomorphic logic into computers, though approximately, is very well handled with Fuzzy Control. Designing system and subsequent changes is neatly done linguistic variables as compared with mathematical modeling. It also come with additional advantages such as faster development cycles, precise control and easy to code and easy to understand programming. It is necessary to note that fuzzy logic and control make way for faster solutions which are in acceptable limits and surely are not comparable with high computational accuracy of classical mathematical techniques, but this compromise is worthy in the sense that these problems cannot be fully represented through traditional models. If tried, resultant would be too complex having high computation cost along with ambiguity in data.

Trajectory planning aimed to realize walking of our biped consists of two parts. One is swing trajectory, phase in which one leg is swinging in air kept in front of the other legs .Other part is body trajectory in which swinging leg complete its motion and is placed on ground, and body start moving.

Inverses Kinematics results in the joint angles corresponding to specified location of end-effectors. Kinematic equations derived for a serial manipulator are a set of polynomials result in multiple configurations for this manipulator of chain type. For parallel manipulators, the specification of location of the end-effectors has simple kinematics equations, yielding the joint parameters relationship [12].

To accomplish this paper we created a database taking reading of numerous people. The data from HMCD, which would be simple, and reliable to capture the angle change of various joints of the body. Reason of using fuzzy logic is obvious and we can heuristically conclude that the gait pattern varies on parameters viz. sex, body weight (heavy or light) and state of mind (relaxed or nervous; happy or sad etc.). It can be argued that simple distance measure will never work precisely while on the other hand fuzzy classifier wills few notches higher.

We aim to create a GAIT cycle for various forms of locomotion like simple walking upgraded to brisk walking and then to running and other forms such as stair climbing. For normal and healthy persons the GAIT cycle generated can be used as an input to realize a biped [1].

### IV. EXPERIMENTAL RESULT

The research shows some useful behavior of human being towards push recovery. The experimentally result using in house development proved that humans have preference for left and right legs analogous to left and right handedness[14]. The strategy is different for left hander or right handed persons. The knee joint is very active in maintaining balance towards push. It gives great insight to develop a reactive controller for a bipedal humanoid robot, to make robot model different from the architecture of plain robot, which robot is helpful in many real world interface like swimming robot or a flying robot. So, It is desirable to use the mass instead of the density so we set density =-1 mass =0.5 density in kilogram/cubic meter and mass in kilometer. When we specified density it ignored mass. The below mentioned table gives details about the various parameters used in our model.

| Parameter | Geometrical Structure | Size(meter) | Mass (Kg) |
|---|---|---|---|
| Head | Sphere | Radius-.04 | 0 |
| Body | Cylinder | Length-0.2, Radius-.008 | 3.8 |
| Thigh(Left& | Cylinder | Length-0.2, | .735 |
| Right) | | Radius-.008 | |
| Leg(Left & Right) | Cylinder | Length-0.2, Radius-.008 | .735 |
| Foot | Cuboids | .04×.02×.02 | .1 |

Table1:- Details of Different Body of our model.

Denavit–Hartenberg parameter or D-H parameter are well known and highly useful concept in robotics. With the help of D-H parameter, next link of the manipulator can fully described with respect to the previous link [20][21].

Right ankle and left ankle of left handed person shows little involvement in balancing. This experiment shows a very different result that one side of human body is more active during push recovery i.e. for left hander person, his right leg joints are more active in push recovery pattern and vice versa. Vital role has played by knee in balancing.

ANALYSIS

The research exhibit the useful behavior for further development of push recovery based model of human being towards push recovery. The data collected by HCMD and research done in our previous paper[3] concluded that left handed or right handed persons use different strategy for push recovery and specially the knee joint is too active during maintain the balance against push. Our developed model controller led for subsequently development of biped reactive controller.

PERFORMANCE MEASURES

While dynamic models used a LIPM, which assume a point mass biped with fixed CoM value. We validate our method on the HOAP-2[2], which have 25 DoF with multibody mass better flexibility to avoid obstacle and less singularity issue as compared to LIPM model

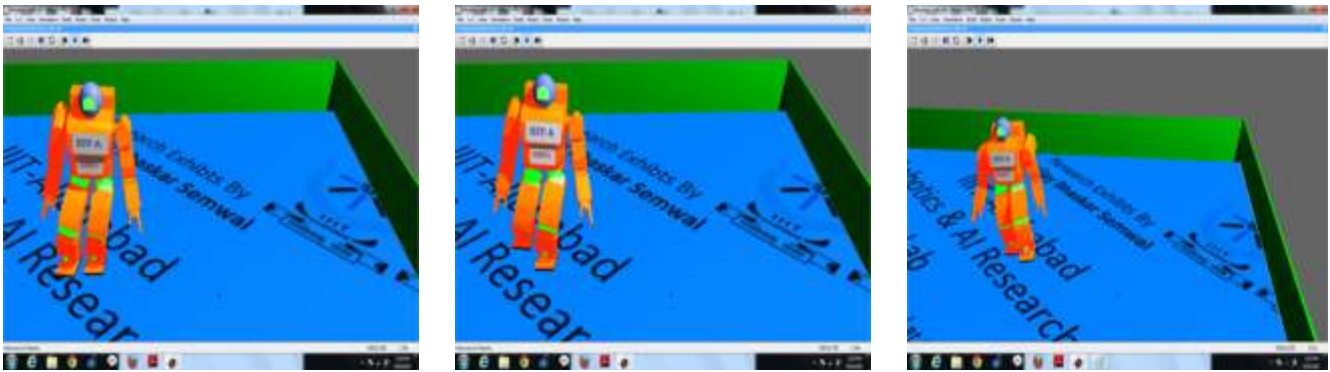

Fig 4: Snap Shot of three different phase of GAIT cycle perform by our model.

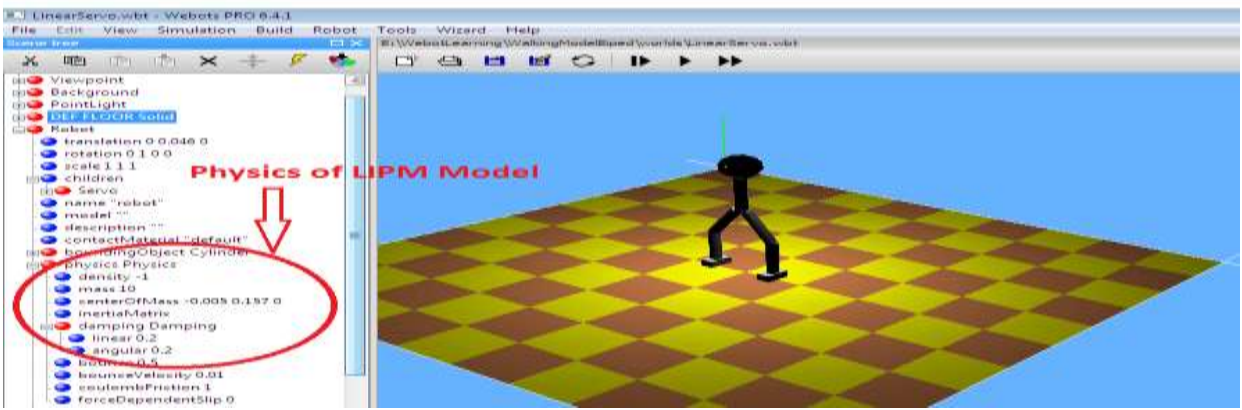

Figure 5: LIBM image from frontal plane in Webots.

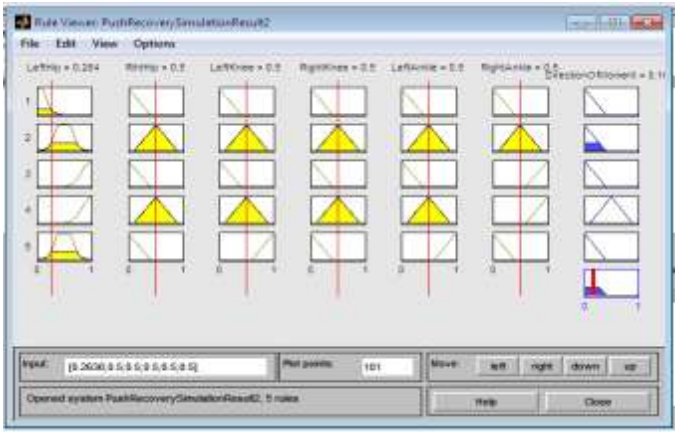

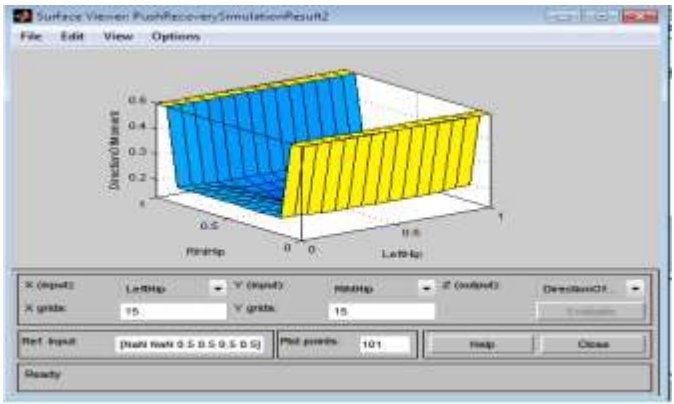

Fig 6: Fuzzy Rule result & Surface View of Hierarchical based Rule

IV. CONCLUSION AND FUTURE DISCUSSION

In direction of push recovery which is highly complex and high dimensional problem the model described is a big step in the direction of generalization of robot architecture. Our simulation model is in accordance with the development of real robot in the same direction keeping HOAP-2 in mind, a biped model humanoid model can be developed having properties such as mass, density, inertia, able to learn from examples, for which we can define physics and in turn it can exhibit some capability for generalization beyond the sample data. Also it has universal approximation property parameters and measurement similar or close to HOAP-2[1]. This was done so as to lower down the complexity of the problem and make it easy to understand the governing reason for push recovery. The model developed in this direction is biped which has legs equipped with number of servos, enough to make it follow ankle and hip strategy as well. The data was gathered and the fuzzy module was developed which showed result with high accuracy. This model and the data gathered can help in further research in this direction of push recovery. To add learning to the model and make it fast in response and remove computation a hierarchal fuzzy model was developed. This is very close to that of actual synthetic data. We have concluded that bipedal push recovery is based on learning and controlled reactions and if we want to impart effective push recovery to a humanoid robot, we will have to understand that it is a software engineering problem rather than hardware controlled problem as envisioned earlier. This will lead to reach on consensus that push recovery is software problem.